\documentclass[conference]{IEEEtran}
\IEEEoverridecommandlockouts
\usepackage{cite}
\usepackage{amsmath,amssymb,amsfonts}
\usepackage{algorithmic}
\usepackage{graphicx}
\usepackage{textcomp}
\usepackage{xcolor}
\usepackage[a4paper, total={184mm,239mm}]{geometry}
\def\BibTeX{{\rm B\kern-.05em{\sc i\kern-.025em b}\kern-.08em
    T\kern-.1667em\lower.7ex\hbox{E}\kern-.125emX}}

\usepackage{subfigure}
\usepackage{multirow}

\begin{document}

\title{CNN-Cap: Effective Convolutional Neural Network Based Capacitance Models for Full-Chip \\Parasitic Extraction}
\author{\IEEEauthorblockN{Dingcheng Yang, Wenjian Yu\IEEEauthorrefmark{1}, Yuanbo Guo, Wenjie Liang}
\IEEEauthorblockA{Dept. Computer Science \& Tech., BNRist, Tsinghua University, Beijing 100084, China\\
Email: ydc19@mails.tsinghua.edu.cn, yu-wj@tsinghua.edu.cn, guoyb17@mails.tsinghua.edu.cn, liang-wj18@tsinghua.org.cn
}}

\maketitle

\begin{abstract}
Accurate capacitance extraction is becoming more important for designing integrated circuits under advanced process technology. The pattern matching based full-chip extraction methodology delivers fast computational speed, but suffers from large error, and tedious efforts on building capacitance models of the increasing structure patterns. 
In this work, we propose an effective method for building
convolutional neural network (CNN) based capacitance models (called CNN-Cap) for two-dimensional (2-D) structures in full-chip capacitance extraction. With a novel grid-based data representation, the proposed method is able to model the pattern with a variable number of conductors, so that largely reduce the number of patterns. Based on the ability of ResNet architecture on capturing spatial information and the proposed training skills, the obtained CNN-Cap exhibits much better performance over the multilayer perception neural network based capacitance model while being more versatile. Extensive experiments on a 55nm and a 15nm process technologies have demonstrated that the error of total capacitance produced with CNN-Cap is always within 1.3\% and the error of produced coupling capacitance is less than 10\% in over 99.5\% probability. CNN-Cap runs more than 4000X faster than  2-D field solver on a GPU server, while it consumes negligible memory compared to the look-up table based capacitance model. 
\end{abstract}
\begin{IEEEkeywords}
full-chip parasitic extraction, capacitance model, pattern matching, machine learning, convolutional neural network
\end{IEEEkeywords}

\section{Introduction}

With the continuous down scaling of process technologies, the interconnect wires in integrated circuit (IC) become smaller, closer to each other, and the integration density  becomes higher. As a result, modeling effects of the interconnect parasitics (mainly including resistance and capacitance) is increasingly important and crucial for guaranteeing the performance of integrated circuits \cite{choudhury1995automatic,lavagno2006eda,yu2021advancements}. 
Nowadays, the parasitic modeling (called parasitic extraction) has become one of the essential steps in the design flow, which is the basis of accurate timing analysis and other performance verification \cite{gong2010parasitic}.

As billions of transistors are placed on a chip, it is challenging to perform the parasitic extraction for tens of billions of interconnect segments.
To solve the full-chip parasitic extraction problem, the pattern matching based techniques are the most widely used, such as in StarRC of Synopsys, QRC of Cadence, and other commercial tools. Another approach of parasitic extraction is based on field solver \cite{nabors1991fastcap,le1992stochastic,yu2013rwcap,wang2005improved,yu2014advanced}, which has the highest accuracy. However, due to excessive computational cost, the field-solver based approach is not suitable for the full-chiap extraction problems. 

The pattern matching approach divides an interconnect layout into small geometries and then calculates the capacitances of each geometry with pre-built empirical formulas or look-up tables of capacitance. A \emph{pattern} refers to the geometries sharing similar topology or formula of capacitance. For a given process technology, a pattern library is pre-characterized by enumerating millions of sample geometries and solving the capacitances of each geometry with field solver. Then, the capacitance values are fitted into formulas associated with geometry, or stored as look-up tables. At the time of extraction, through pattern matching the capacitances of input geometries can be calculated quickly and the capacitances of nets are obtained by assembling the capacitances of these geometries. 

However, the pattern matching approach is facing the following challenges. 1). The number of patterns increases with the advancement of process technology, and it becomes difficult to make the patterns covering all possible interconnect typologies in real design. 2). The look-up table based approach storing capacitance values of sample geometries consumes enormous or unaffordable memory space for achieving good accuracy, while the error of empirical formulas increases as well. 3). the pattern matching approach needs a large number of capacitance values produced by field solver, which often takes longer time for a process technology as the metal/dielectric configuration becomes complex \cite{yu2021advancements}.
So, there is a strong need for new capacitance modeling technique to improve the performance of pattern matching based method.

Although the process of fitting capacitance formulas for a structure pattern is a regression problem and deep neural networks (DNNs) have achieved notable successes in many classification and regression problems in recent years \cite{lecun2015deep}, only a few of published work are about employing DNN in the area of parasitic extraction  \cite{sen2006neural,yao2016machine,kasai2019neural,shook2020mlparest,li2020layout}. Moreover, most of them either deal with the numerical computing in the field-solver approach \cite{yao2016machine}, or do not involve the capacitance calculation for a given interconnect geometry \cite{shook2020mlparest}. The most relevant work to the capacitance modeling or calculation is \cite{kasai2019neural}, where a neural network based method is presented for several structure patterns in three-dimensional (3-D) ICs. Nevertheless, it only considers single-dielectric structures with simple multilayer perception (MLP) neural networks, and the demonstrated error on total capacitance can be larger than $10\%$ \cite{kasai2019neural}. The practicality and  effectiveness of the technique in \cite{kasai2019neural} is obviously not good. 
Instead of directly calculating capacitances, an MLP neural network based approach was proposed to improve the pattern matching based capacitance extraction through automatic pattern classification and capacitance formula building \cite{li2020layout}.

In this work, we propose a convolutional neural network (CNN) based capacitance modeling method for improving the pattern matching based extraction methodology. 
The major contributions are as follows.
\begin{itemize}
    \item A grid-based data representation and the corresponding DNN modeling approach are proposed for 2-D structure pattern with a variable number of conductors. It separates the tasks of calculating total capacitance and coupling capacitance, so as to potentially reduce the difficulty of training accurate model for capacitance extraction. Moreover, allowing a pattern to include a variable number of conductors largely reduces the number of patterns and therefore the efforts on building capacitance models.
    \item A CNN model called CNN-Cap, which is derived from the ResNet architecture and inherits the ability of capturing spatial information, is proposed for predicting the capacitances of the pattern with a variable number of conductors. A training approach including a loss function for more accurate coupling capacitance is proposed to make CNN-Cap suitable for modeling the pattern capacitances.
\end{itemize}

Extensive experiments with two process technologies demonstrate that the proposed CNN-Cap has much better accuracy on capacitance calculation than the counterpart model based on MLP neural network. For all the tested pattern structures, CNN-Cap is able to predict all total capacitances with less than \textbf{1.3\%} error, and \textbf{over 99.5\%} of coupling capacitances with error less than 10\%.  
The sensitivity of the model's performance to the size of training data is also studied, which shows that CNN-Cap performs well with less training effort. Finally, CNN-Cap runs \textbf{4693X} faster than 2-D field solver on a GPU server, while it consumes negligible memory compared to the look-up table based capacitance model.

\section{Background}
\subsection{Full-Chip Capacitance Extraction Based on Pattern Matching and 2.5-D Extraction Method}

For full-chip capacitance extraction, directly using the 3-D field solver is infeasible due to its excessive cost of memory and CPU time. To obtain good tradeoff between accuracy and efficiency,  the 2.5-D extraction method with pattern matching technique is widely used. The pattern matching based extraction methodology include three major modules: 1) pattern generation,  2) capacitance model building, and 3) layout capacitance extraction \cite{lavagno2006eda,yu2014advanced,li2020layout}. It is illustrated as Fig. 1. For a given process technology, the work of 1) pattern generation and 2) capacitance model building are carried out just once.
\begin{figure}[h]
\setlength{\abovecaptionskip}{1pt}
  \centering
    \includegraphics[width=3.6in]{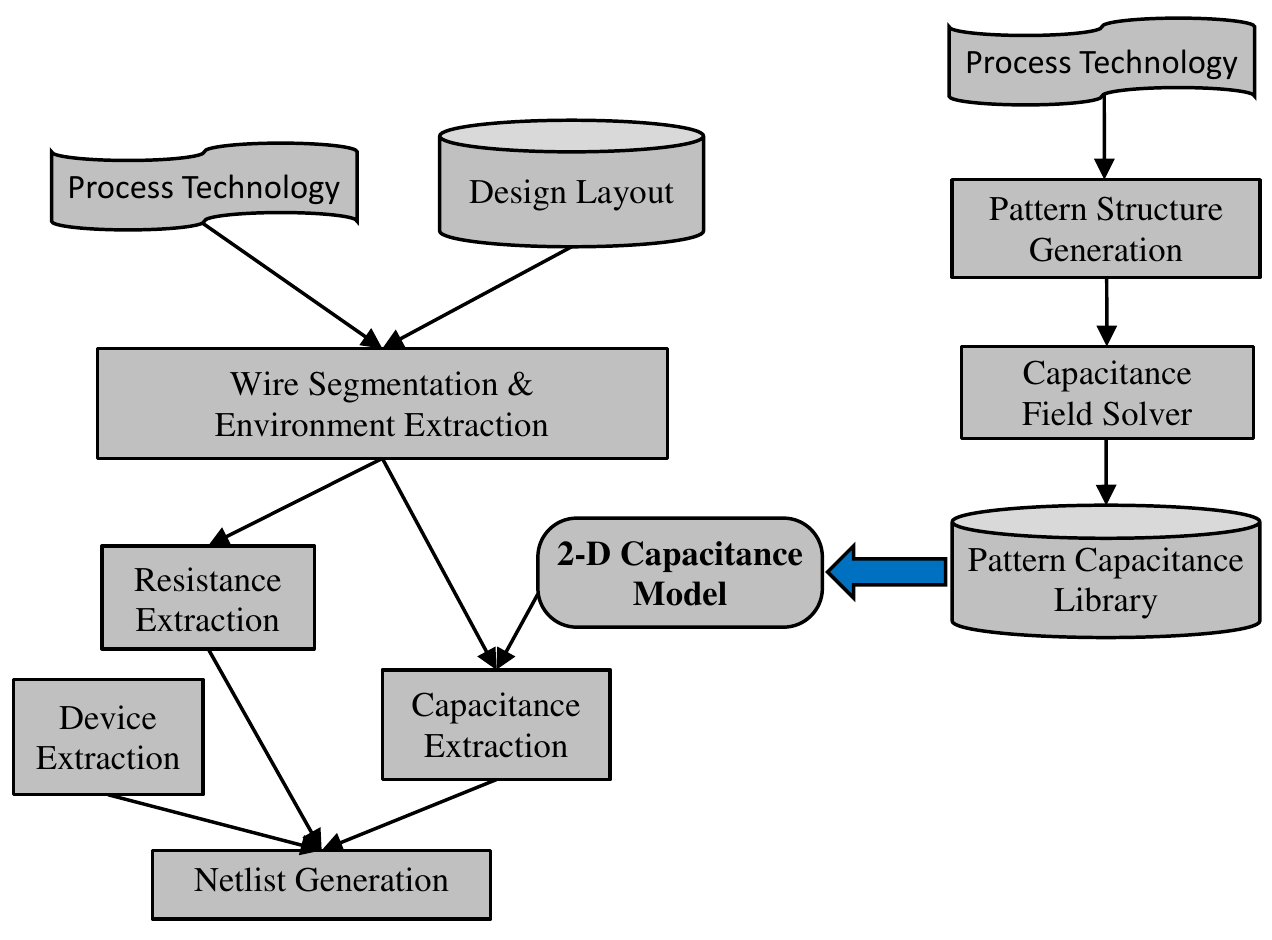}
  \caption{The pattern matching based full-chip parasitic extraction \cite{yu2021advancements}.}
    \label{fig:patternmatching}
\end{figure}

The 2.5-D extraction method refers to the method considers 3-D geometric effects on capacitance with 2-D interconnect patterns through sweeping 3-D geometry of interconnects in two perpendicular directions \cite{yu2005capacitance,cong1997analysis}. It is adopted by the OpenROAD project for parasitic extraction \cite{openRCX}.
During the layout capacitance extraction,  extraction windows are generated along the interconnect line (called \emph{master net} or \emph{master conductor}) whose capacitances are of concern. Each window 
 includes a segment of the master net and its neighbor conductors (called \emph{environmental conductors}). The techniques of 2.5-D extraction and pattern matching are employed to calculate the capacitances among the conductors in the window. 
Take the structure shown in Fig. 2 as an example, where a wire with name $m2$ crosses over a wire named $m1$. Along direction A, a 2-D cross-section view is shown in the middle of Fig. 2. Along direction B, the other 2-D cross section is shown to the right. If the capacitances in the two 2-D cross-section views are known, we can approximately compute the capacitance between $m1$ and $m2$ with the 3-D effects taken into consideration. Suppose the 2-D capacitance between $m1$ and $m2$ in the view along A is
\begin{equation}
C_A= C_{1f1}+C_{1o}+C_{1f2},
\end{equation}
where $C_{1f1}$ and $C_{1f2}$ are two fringe capacitances, and $C_{1o}$ is the overlapping capacitance. Similarly,
\begin{equation}
C_B= C_{2f1}+C_{2o}+C_{2f2},
\end{equation}
for the view along B. Then,
\begin{equation}
C_{m1,m2}= C_A w_1+(C_B-C_{2o})w_2,
\end{equation}
where $w_1$ and $w_2$ are widths of wires $m1$ and $m2$, respectively. 
\begin{figure}[b]
 \setlength{\abovecaptionskip}{1pt}
  \centering
    \includegraphics[width=3.3in]{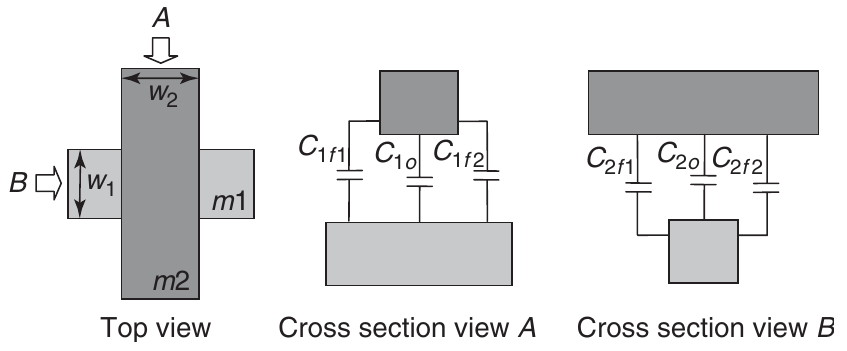}
  \caption{2.5-D capacitance calculation for a crossover structure \cite{yu2005capacitance}.}
    \label{fig:2p5Dextraction}
\end{figure}

\begin{figure}[t]
\setlength{\abovecaptionskip}{1pt}
  \centering
    \subfigure[]
    {\includegraphics[width=2.7in]{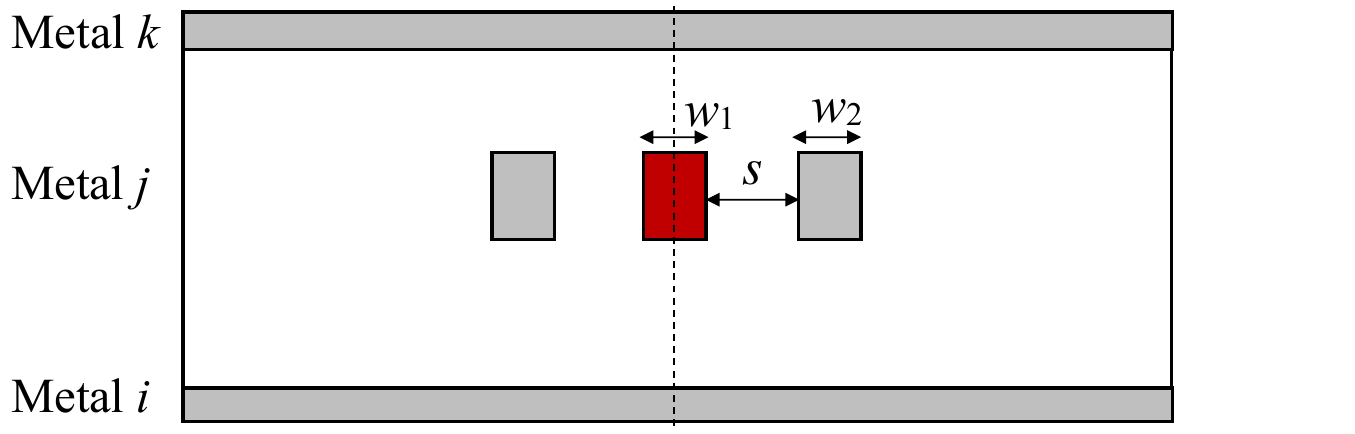}\label{fig:3L2G}}
    \subfigure[]
    {\includegraphics[width=2.7in]{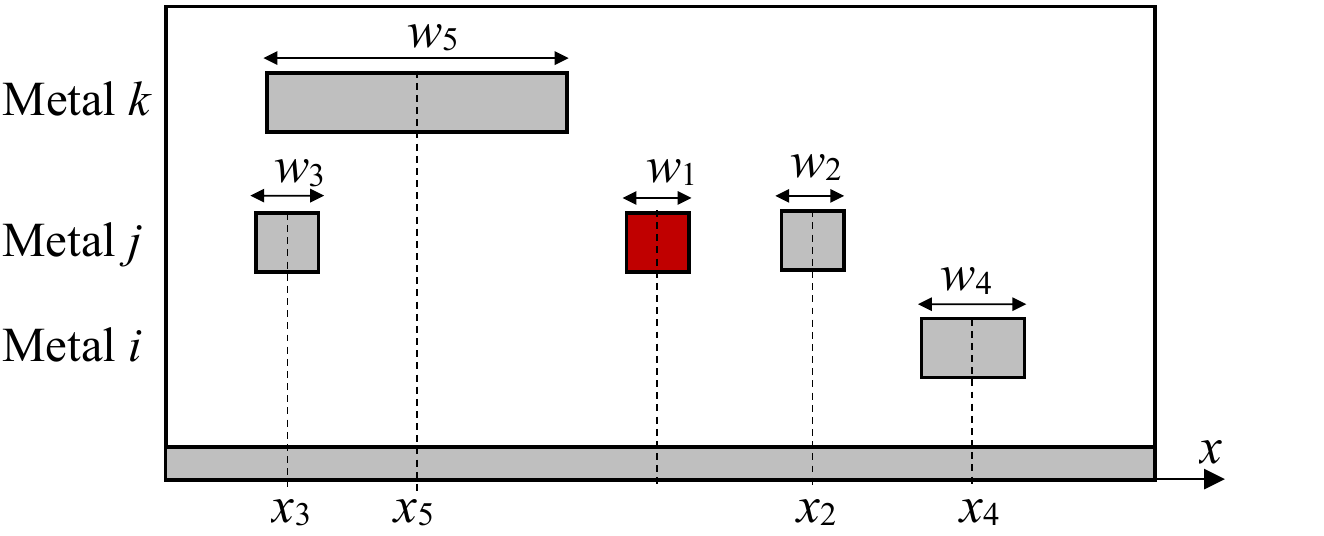}\label{fig:pattern2}}  
  \caption{Two typical 2-D interconnect  structure patterns (multi-dielectric environment is not drawn). (a) Pattern-A:  ``sandwich'' structure with three parallel wires. (b) Pattern-B: three metal layers with a fixed number of conductors.}
    \label{fig:pattern}
\end{figure}

With this method, one can only consider the capacitance models for the 2-D cross-section structures. These structures are regarded as the instances of pre-defined 2-D patterns. Two kinds of typical structure patterns are shown in Fig. \ref{fig:pattern}. Pattern-A shown in Fig. \ref{fig:3L2G} is a ``sandwich'' structure including two big-plane conductors and three parallel wires in between. The red block denotes the master conductor, and the whole structure is usually left-right symmetrical. Pattern-B's structure is more general than A, where the conductors on the top and down metal layers are not required to be a big plane across the extraction window. On the layer where the master conductor lies, there are two conductors on the left and right side of the master respectively. At the very bottom, there is an extra big-plane conductor representing the substrate ground. 
Although not shown in Fig. \ref{fig:pattern}, the multi-dielectric environment is considered in these structure.

Usually in a window, only the conductors located on the nearest metal layers above and blow the master conductor are considered. Due to the proximity  effect of electrostatic field, this brings little error to the capacitances of the master. Therefore, most patterns are defined  by three metal layers (containing master conductor and its neighbour conductors) in a given process technology, and the numbers of conductors on each layer. 
A capacitance model for a pattern is built through computing the capacitances of a lot of instance structures with field solver and then building capacitance models  with look-up tables or curve fitting techniques.

\subsection{Neural Network Based Capacitance Extraction}
\begin{figure}[t]
\setlength{\abovecaptionskip}{1pt}
  \centering
    \includegraphics[width=3in]{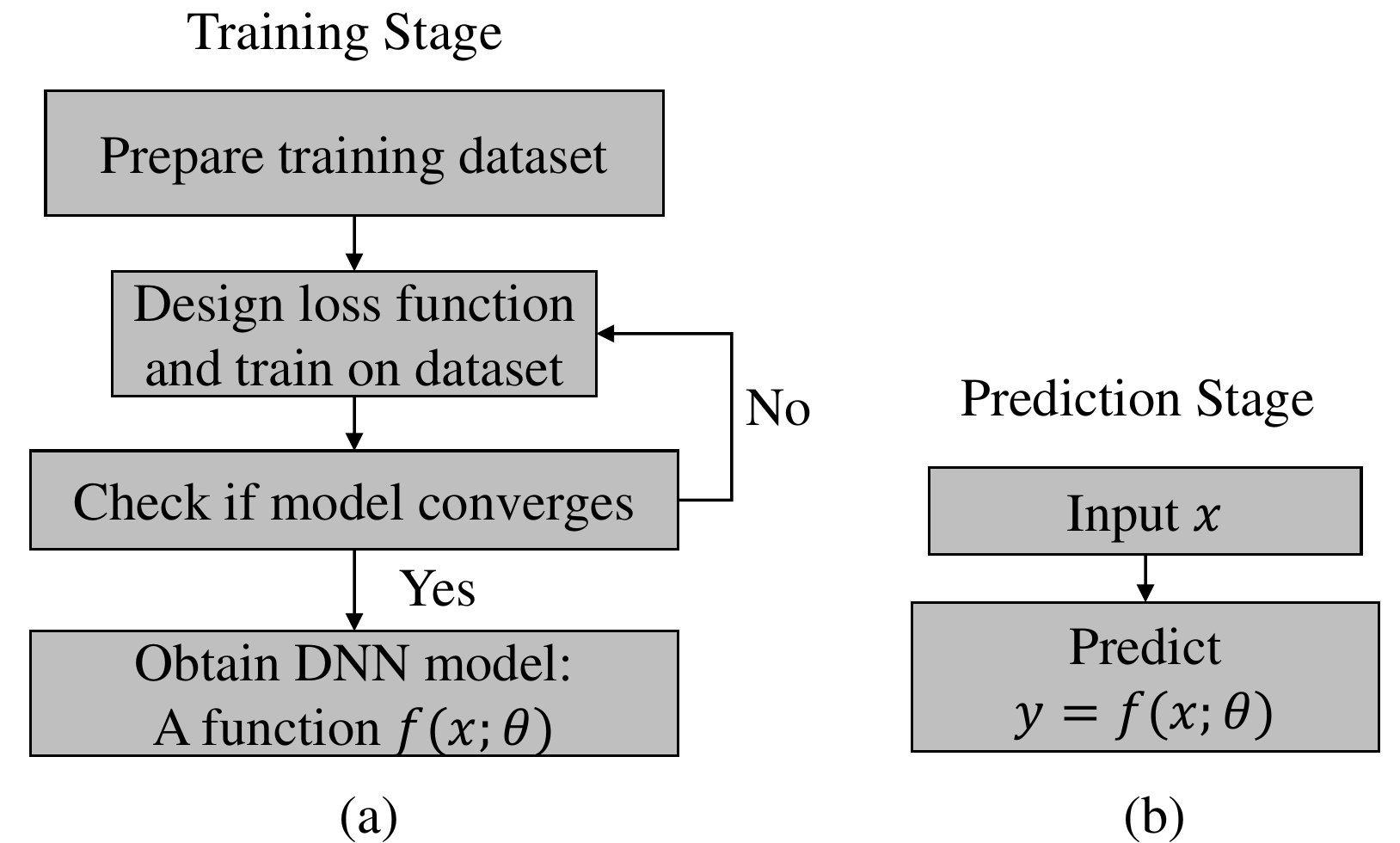}
  \caption{The work flow of a DNN: (a) training stage, (b) prediction stage.}
    \label{fig:pipeline}
\end{figure}

The neural network with multiple hidden layer of neurons is often refers to as deep neural network (DNN).
    Let $\boldsymbol{x} \in \mathbb{R}^n$ be the input data and $\boldsymbol{y} \in \mathbb{R}^m$ be the output data, then, the DNN can be viewed as a function $f(\boldsymbol{x}; \boldsymbol{\theta}): \mathbb{R}^n \rightarrow \mathbb{R}^m$ with parameters $\boldsymbol{\theta}$. The form of the function $f$ determines the type of DNN. The training of a DNN and the prediction with a trained DNN are illustrated in Fig. \ref{fig:pipeline}. During the training, a large amount of input data along with the corresponding outputs (called labels) are fed to the network. The network parameters $\theta$ are optimized to minimize the difference between $f(\boldsymbol{x}; \boldsymbol{\theta})$ and the corresponding labels. When the difference is sufficiently small, the network is trained to become a good regressor, which can be used to do prediction (see Fig. \ref{fig:pipeline}(b)). 

\begin{figure}[h]
\setlength{\abovecaptionskip}{1pt}
  \centering
    \includegraphics[width=2.3in]{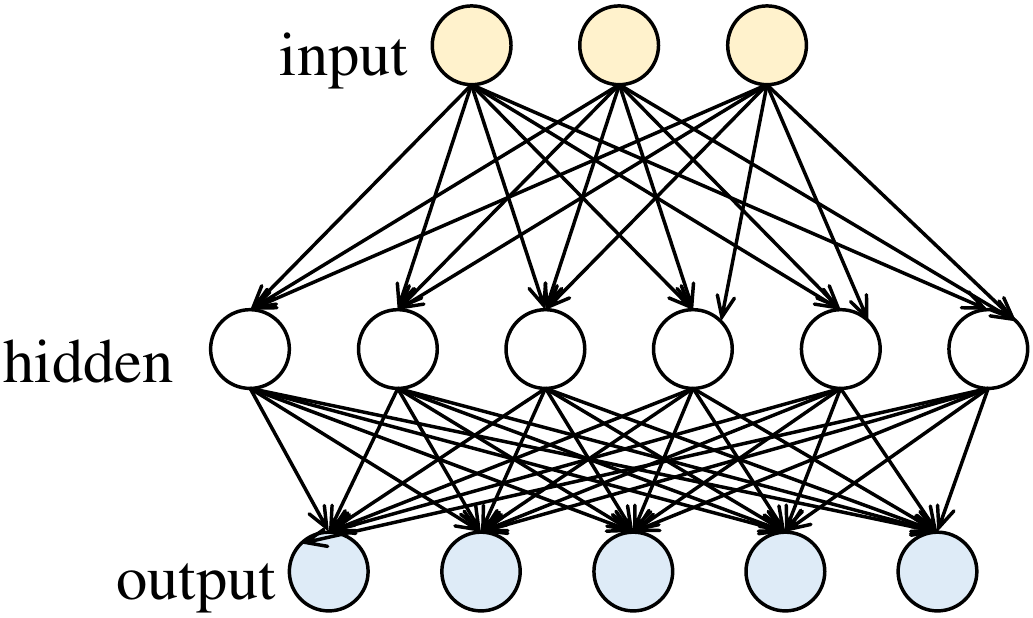}
  \caption{An example of MLP network. The yellow, white, and blue circles stand for neurons in the input, hidden, and output layers.}
    \label{fig:mlp}
\end{figure}

The MLP neural network is a simple yet popular neural network, as shown in Fig. \ref{fig:mlp} where only one hidden layer is drawn. Suppose there are $n_l$ hidden layers. Let $\boldsymbol{h}^{(i)}$ be the variables residing at $i$-th hidden layer's neurons. Those on the input layer and output layer can be denoted by $\boldsymbol{h}^{(0)}$ and $\boldsymbol{h}^{(n_l+1)}$, respectively. The input data elements are assigned to each neuron in the input layer and feed-forward to the next layer iteratively until they reach the output layer. For the $i$-th layer, it means $\boldsymbol{h}^{(i)}=g_i(\boldsymbol{h}^{(i-1)})$ where $g_i$ is a function to represent the feed-forward process. In general, $g_i$ is a compound function of a nonlinear activation function and a linear function including weight parameters.

The MLP network has been studied in capacitance extraction of several 3-D structures \cite{kasai2019neural}. The idea is straightforward and can be applied to build the capacitance model of the 2-D patterns. For Pattern-A, there are just three width/spacing parameters for describing the structure: $w_1,w_2$ and $s$ (see Fig. \ref{fig:3L2G}). For the example of Pattern-B (see \ref{fig:pattern2}), the parameters are widths  $w_1,w_2,w_3,w_4, w_5$, and location coordinates $x_2,x_3,x_4, x_5$.
This makes the regression model well defined; the input $\boldsymbol{x}$  is just a vector including these geometry parameters, and the output $\boldsymbol{y}$ is a vector including the total capacitance and coupling capacitances.
With a lot of sample structures for a pattern and corresponding capacitances results from field solver, the MLP based model can be trained. 

Besides training approach and model architecture, different loss function also affects the difficulty of optimization process and the performance of trained model. Usually the mean square error (MSE) is employed as the loss function for minimization:
\begin{align}
    MSE=\frac{1}{N}\sum_{i=1}^N \|f(\boldsymbol{x}^{(i)};\boldsymbol{\theta})-\boldsymbol{y}^{(i)}\|^2,
\end{align}
where $N$ is the number of training examples, $\boldsymbol{x}^{(i)}$ indicates the $i$-th example, and $\boldsymbol{y}^{(i)}$ is the corresponding label. 

\section{Building Effective Capacitance Models Based on Convolutional Neural Network}
In this section, we first propose the idea of leveraging CNN to improve the capacitance modeling of structure patterns. Then, we present a grid-based data representation, the CNN architecture and the training approach in details. Finally, we present the data generation approach and other details.

\subsection{The Basic Idea}
In all existing methods, a pattern in the 2.5-D extraction flow is determined by a combination of metal layers and the numbers of conductors on these layers. And, the geometric parameters characterizing a pattern increases with the number of conductors in the pattern. For a pattern with a large quantity of  parameters, building an accurate model with either traditional approaches or the MLP based approach becomes difficult. To overcome this issue, we view the 2-D cross-section structure as a kind of image, and try to characterize capacitance related information  within it with the  convolutional neural network which performs very well in image processing applications. So, in this way, we can allow a pattern including a variable number of conductors and thus largely reduce the number of patterns.

\begin{figure}[h]
\setlength{\abovecaptionskip}{1pt}
  \centering
\includegraphics[width=2.6in]{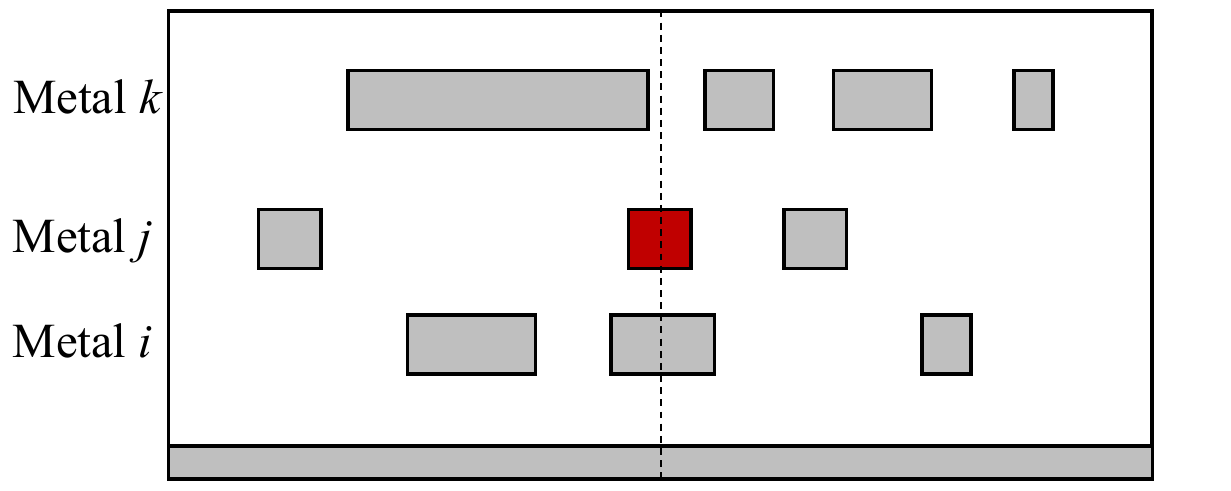}  
  \caption{A structure of three metal layers with a variable number of conductors (dielectric environment not depicted). The structure is referred to as Pattern-C. }
    \label{fig:newpattern}
\end{figure}

This considered pattern is illustrated by Fig. \ref{fig:newpattern}. Its structure is more general than that of Pattern-B. The master conductor is still at the center of the middle layer. Although the conductors can lie on more than three metal layers, we assume they are just on three metal layers (with index $i$, $j$ and $k$) in the rest of this paper. The proposed method can be easily extended to the patterns with more than three metal layers. The structure in Fig. \ref{fig:newpattern} is referred to as Pattern-C.

\subsection{Grid-Based Data Representation}

\begin{figure}[h]
\setlength{\abovecaptionskip}{1pt}
  \centering
    \includegraphics[width=3.54in,height=0.92in]{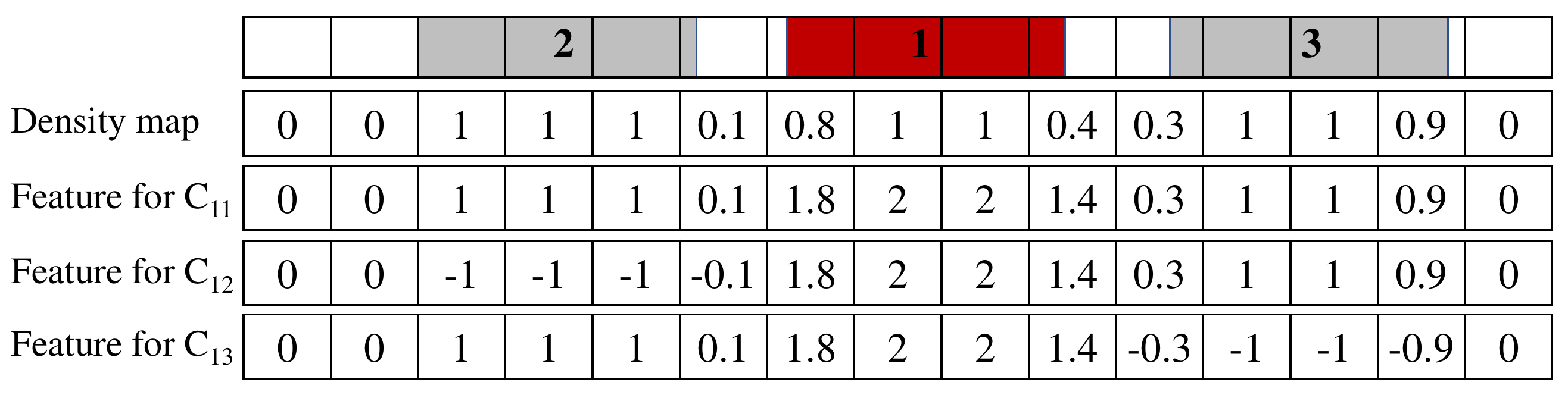}
  \caption{Grid-based data representation for a metal layer in a pattern. A case with three conductors is shown in the first row. The following rows are the density map and the data representations for the calculation of total capacitance and coupling capacitances.}
    \label{fig:input}
\end{figure}

Inspired by the idea of encoding the layout of interconnect wires with density-based features in the design for manufacture (DFM) research \cite{wen2014fuzzy}, we propose to evenly divide the width of extraction window into $L$ grid cells. Thus, the conductor placement of a metal layer can be depicted by an   $L$-dimensional vector. The value of vector element is the density, i.e. the ratio of conductor occupying the grid cell. See the example in Fig. \ref{fig:input}, where the horizontal placement of three conductors (labeled 1, 2 and 3), the grids, and the corresponding density map are shown.
For a Pattern-C structure, the conductor placement can be described with three  $L$-dimensional vectors. Notice this representation is not ambiguous if we guarantee that the grid size is less than the minimum spacing between wires on the metal layer.

This grid-based representation cannot identify the master conductor. So, we must encode more information regarding the  pattern into it.
Suppose the structure includes $n_c$ conductors. Our problem is to extract one total capacitance, and $n_c-1$ coupling capacitances. Consider the density-based representation for one metal layer: $\boldsymbol{d} \in \mathbb{R}^L$.
We modify it to obtain an $L$-dimensional vector $\boldsymbol{x}$ for a sub-problem of calculating total capacitance. The scheme of  the modification is that, if the master conductor covers the $i$-th grid, we have $x_i=d_i+1$. The obtained feature vector $\boldsymbol{x}$ is shown as the third row in Fig. \ref{fig:input}. 
For calculating the coupling capacitance between the master and an environmental conductor, a unique feature vector is also generated by modifying $\boldsymbol{d}$. Besides adding 1 to the elements corresponding the cells overlapped with the master, another modification is applied to identifying the environmental conductor. If the environmental conductor covers the $i$-th grid, we make that $x_i=-d_i$. Now, for calculating different coupling capacitance there is a different feature vector for data representation. In  Fig. \ref{fig:input}, the last two rows show the feature vectors for the sub-problems of calculating coupling capacitances $C_{12}$ and $C_{13}$.

\begin{figure}[h]
\setlength{\abovecaptionskip}{1pt}
  \centering
    \includegraphics[width=3.54in]{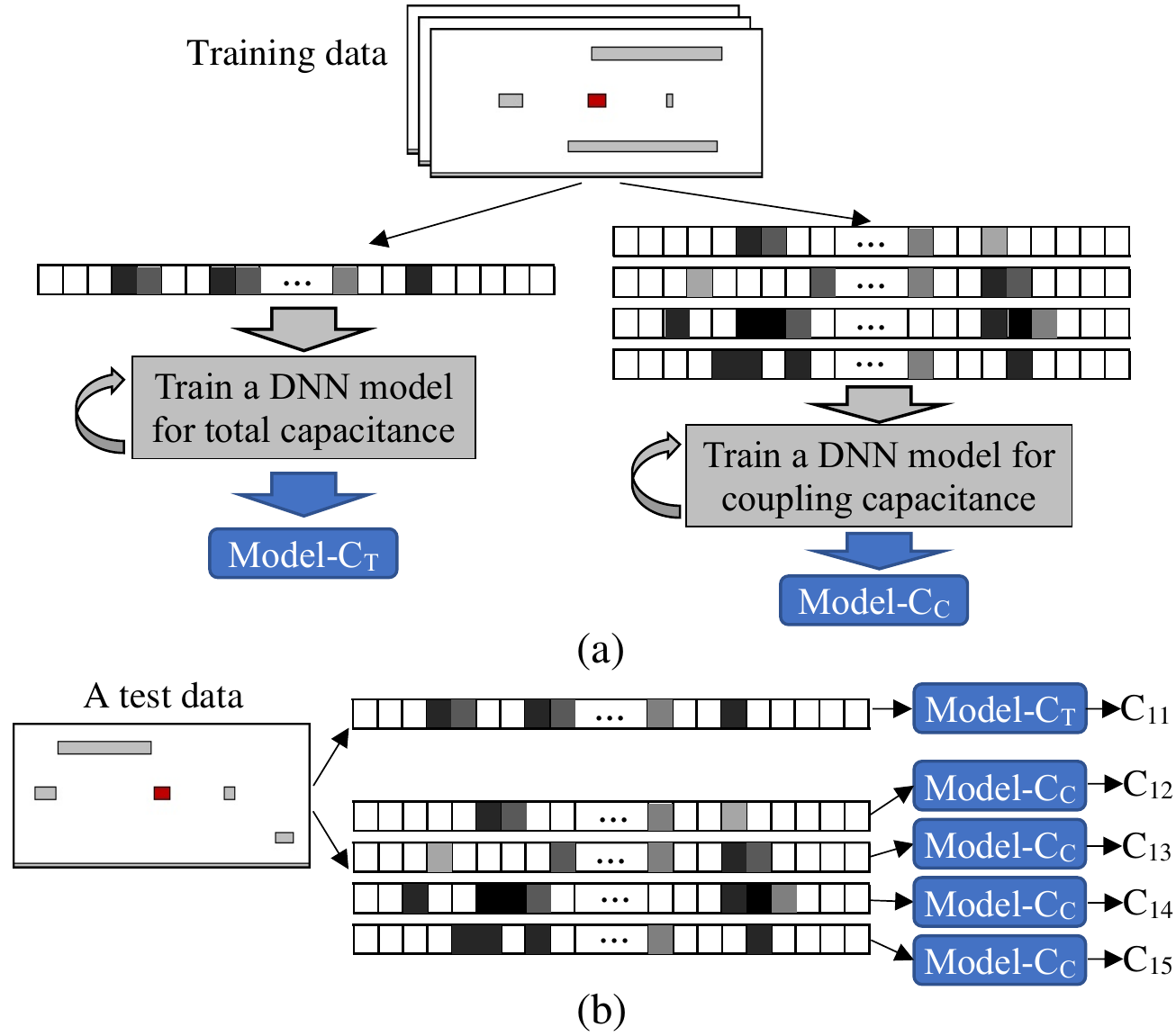}
  \caption{The proposed method for building capacitance models for Pattern-C. (a) The training stage. (b) The prediction stage.}
    \label{fig:approach}
\end{figure}

Because the values of total capacitance and coupling capacitance can differ for several orders of magnitude, the accuracy demand for them is usually different. In this work, we distinguish the problems of calculating total capacitance and of coupling capacitances, and trains two models for them respectively for a given Pattern-C. This ensures the overall accuracy of capacitance extraction. With this idea and the proposed grid-based data representation, our method for building the capacitance models can be depicted as Fig. \ref{fig:approach}. Notice that a sample structure is converted to $n_c$ data inputted to the two models in the training stage.
And, in the prediction stage, the trained model Model-C$_\mathrm{C}$ for coupling capacitance will be evaluated for $n_c-1$ times to output the $n_c-1$ coupling capacitances.

\subsection{CNN Architecture and Training Approach}
Because the CNN is able to capture the spatial information in the data, it is expected to perform better than MLP neural network for the problem of capacitance calculation. So, we develop the CNN base models (called CNN-Cap) for predicting the capacitances of Pattern-C structures. There are two models with same architecture used for total capacitance and coupling capacitance, respectively.

\begin{figure}[b]
\setlength{\abovecaptionskip}{1pt}
 \centering
    \includegraphics[width=3.4in]{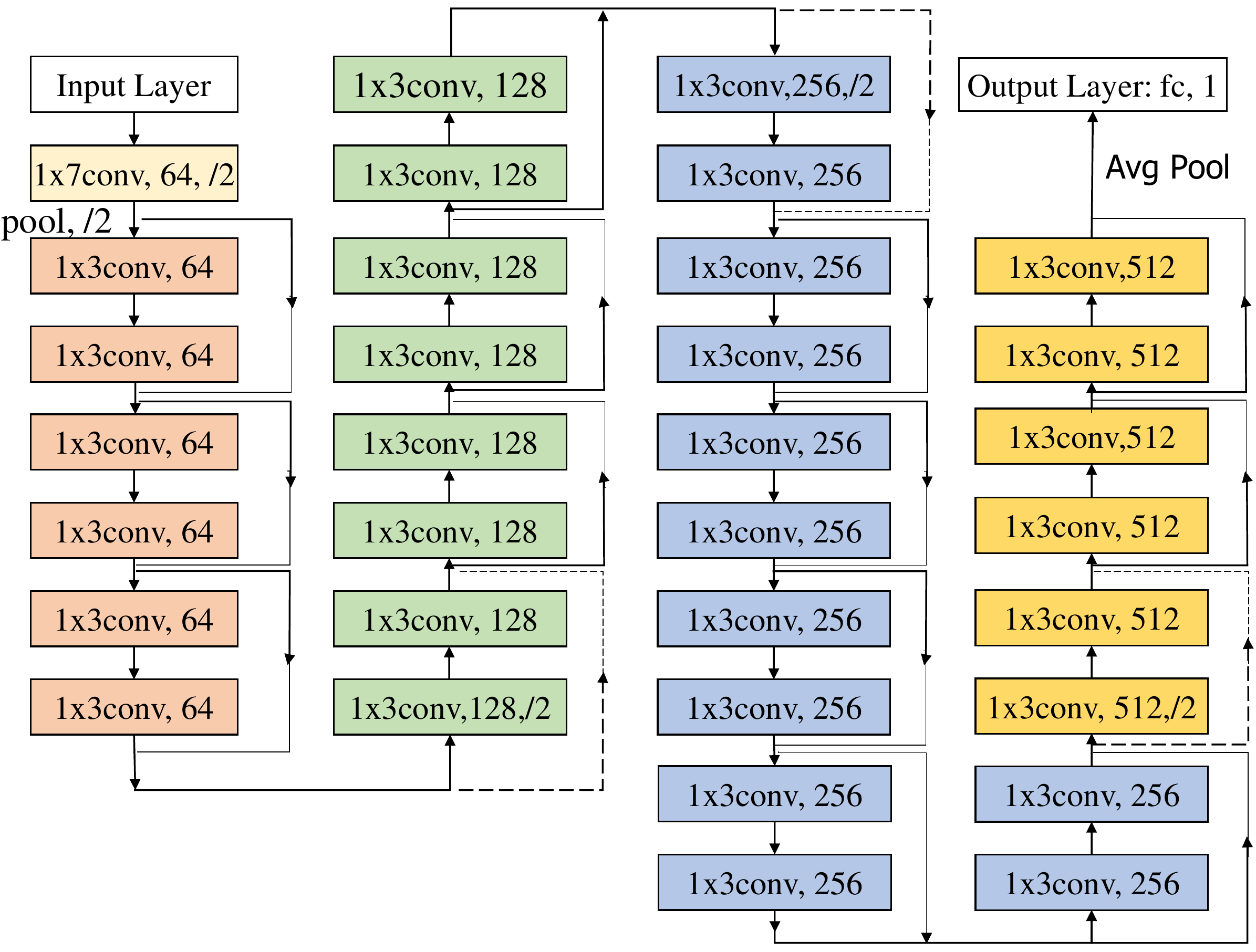}
 \caption{The architecture of proposed CNN-Cap.}
    \label{fig:resnet}
\end{figure}

The architecture of CNN-Cap is shown in Fig. \ref{fig:resnet}, which is derived from the famous ResNet architecture \cite{he2016deep}. In our problem, the input data (pattern structure with extraction information) is vectors, instead of the matrices in image related problem. So, we just use 1-D convolutional layer (shown as colored block in Fig. \ref{fig:resnet}) in CNN-Cap. Notice that the number of metal layers in the pattern is like the concept of ``channel'' of image data. 
A batch normalization and a ReLU layer are inserted after every convolutional layer. The input data will reach the last convolutional layer through stacked convolutional layers.  Then, it will be pooled into a fixed-length vector by an average pooling layer. Finally, the fixed-length vector will be fed to the fully-connected layer with an output dimension of one to predict capacitance. In  Fig. \ref{fig:resnet}, the convolutional layer captioned ``1x3 conv, 64'' means it has 64 channels and a kernel size of 3. ``/2'' means to halve the input length (by a 1-D pooling layer or a 1-D convolutional layer with a stride of 2). The convolutional layers with the same color mean they have the same input length. Whenever a 1-D convolutional layer halves the input length, the input channel will be doubled. The key point of CNN-Cap is the shortcut connection, which takes a shallower vector $\boldsymbol{h}_1$ and a deeper vector $\boldsymbol{h}_2$ as input and takes $\mathcal{F}(\boldsymbol{h}_1)+\boldsymbol{h}_2$ as output. The function $\mathcal{F}$ is an identity mapping in most cases (solid line). The dotted shortcut takes a 1-D 1x1 convolutional layer (equivalent to linear projection) as the mapping function when the shape of two input vectors is mismatched. Without the shortcut connection, a deeper DNN model will have both higher testing error and training error. The higher training error means that it is difficult to train a deep plain model. The shortcut connection can preserve the low-level features all the time, which makes training easier. Thanks to the advanced modules, i.e., 1-D convolutional layer and shortcut connect, our CNN-Cap has better learning ability than the MLP neural networks. This will be revealed with experimental results.

We have tested three architectures derived from ResNet-18, ResNet-34, and ResNet-50. We found out that derived from ResNet-34 is consistently better than that from ResNet-18, and performs similarly to that from ResNet-50. So, the ResNet-34 derived architecture is  chosen in this work.

For training CNN-Cap, the stochastic gradient descent (SGD) optimizer and the Adam  optimizer \cite{kingma2014adam} are considered.  Our experimental results show the Adam optimizer makes convergence easier to reach and the trained model perform better. So, it is employed in this work. 

The approach of grid search is used to find the appropriate learning rate and batch size, which affect performance of the obtained CNN-Cap model. The value range of learning rate is from $10^{-5}$ to $10^{-2}$, and the batch size is enumerated in $\{16, 32, 64, 128\}$.

For the cost function, in addition to the MSE loss function (4), we have tried the following loss function.
\begin{align}
    MSRE=\frac{1}{N} \sum_{i=1}^N (1-\frac{f(\boldsymbol{x}^{(i)};\boldsymbol{\theta})}{\boldsymbol{y}^{(i)}})^2 ~,
\end{align}
where $N$ is the number of training data,  $\boldsymbol{x}^{(i)}$ indicates the $i$-th input data, and $\boldsymbol{y}^{(i)}$ is the corresponding label. The division of (5) stands for an element-wise operation. This MSRE loss function includes the relative error, so that the optimization could possibly attain a better accuracy. However, it may bring difficulty to the convergence of training process. We have done extensive experiments, and found out that the loss function MSRE is better than MSE for the task of training the Model-C$_{\mathrm{C}}$ for coupling capacitance.

\subsection{Dataset Generation and Other Discussion}

In this work, a data is a 2-D structure of cross-section view with the proposed data representation, and the label includes the capacitances computed with a field solver.  

For a given process technology and a specified triple of ($i$, $j$, $k$) of metal layer indices, we can generate random sample structures of Pattern-C. We ensure that the data complies with the rules of minimum width and minimum spacing of the process technology. 
As the structure is a cross-section view along the interconnect line, we  let the width of master conductor 
be a smaller value with larger probability and not exceed 10 times of the minimum width. Similarly, the random data can be generated for the conventional patterns, like Pattern-A and Pattern-B. Notice we can also build the CNN-Cap model for a conventional pattern.

The whole width of the extraction window is determined with a simulation experiment to detect how far if an environmental conductor is away from the master conductor the coupling capacitance between them decreases to 1\% of the total capacitance. 

\begin{figure}[h]
\setlength{\abovecaptionskip}{1pt}
  \centering
    \subfigure[]
    {\includegraphics[width=1.7in]{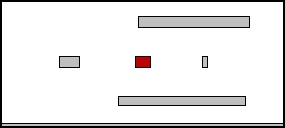}\label{fig:exp0}}  
    \subfigure[]
    {\includegraphics[width=1.7in]{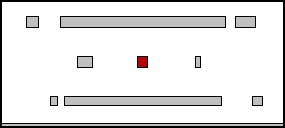}\label{fig:exp1}}  
  \caption{Two randomly generated data for: (a) Pattern-B and (b) Pattern-C. The dielectric environment is not depicted.}
    \label{fig:exp}
\end{figure}

Fig. \ref{fig:exp} shows the geometries of the conductors in two data generated with our approach. Fig. \ref{fig:exp0} stands for a data for Pattern-B and Fig. \ref{fig:exp1} is for Pattern-C.

The existing approach with MLP neural network cannot handle Pattern-C, because the structure involves a variable number of conductors. Instead, we can combine the MLP with the grid-based data representation. It means we connect the three vectors to a form a long vector and then input it to the MLP network. However, this hybrid model cannot have high accuracy, because the spatial information in the grid-based representation is lost. 
The comparison experiments in Section IV.B validate this.

\section{Experimental Results}
2-D pattern structures are generated following two process technologies. One is a  55nm process technology  from industry, while the other is the 15nm process technology in FreePDK15 \cite{bhanushali2015freepdk15,FreePDK15}. The width of structure (extraction window) is set to $56\tilde{w}_{min}$, where $\tilde{w}_{min}$ is the minimum width of the metal layer where the master conductor resides. For the structures, we obtain the capacitance results (as labels of data) with the golden-standard capacitance solver Raphael \cite{raphael}. All tested DNN models are implemented with PyTorch, and all experiments are carried out on a Linux server with 2 Intel Xeon Silver 4214 CPU at 2.2GHz and 8 Nvidia RTX2080Ti GPUs.

The geometric parameters for the metal layers in the two process technologies are listed in Table \ref{table:itf}.
\begin{table}[h]
\setlength{\abovecaptionskip}{1pt}
\centering
\caption{The Geometric Parameters of the Metal Layers in the Two Tested Process Technologies (in Unit of $\mu m$)}
\label{table:itf}
\begin{tabular}{@{~~}c@{}c@{~~}c@{~~}c@{~~}c@{~~}c@{~~}c@{~~}c@{~~}c@{~~}}
\hline
	\multirow{2}{*}{Layer index}& & \multicolumn{3}{c}{55nm Tech.} & & \multicolumn{3}{c}{15nm Tech.}\\
	\cline{3-5} \cline{7-9}

     & & thickness & $w_{min}$ & $s_{min}$ & & thickness & $w_{min}$ & $s_{min}$ \\
      \hline
      1 & & 0.1 & 0.054 & 0.108 & & 0.06 & 0.028 & 0.036 \\
      \hline
      2 & & 0.16 & 0.081 & 0.08 & & 0.06 & 0.028 & 0.036 \\
      \hline
      3 & & 0.2 & 0.09 & 0.09 & & 0.06 & 0.028 & 0.036 \\
      \hline
      4 & & 0.2 & 0.09 & 0.09 & & 0.06 & 0.028 & 0.036 \\
      \hline
      5 & & 0.2 & 0.09 & 0.09 & & 0.06 & 0.028 & 0.036 \\
      \hline
      6 & & 0.2 & 0.09 & 0.09 & & 0.13 & 0.056 & 0.056 \\
      \hline
      7 & & 0.85 & 0.36 & 0.36 & & 0.13 & 0.056 & 0.056 \\
      \hline
      8 & & - & - & - & & 0.13 & 0.056 & 0.056  \\
      \hline
      9 & &- &- & - & & 0.13 & 0.056 & 0.056  \\
      \hline
      10 & &- &- &- & & 0.13 & 0.056 & 0.056  \\
    \hline
\end{tabular}
\end{table}

For the two technologies, we randomly choose some three-metal-layer combinations to generate the 2-D patterns. For each pattern, 
we generated $50000$ sample structures with the approach in Section III.D to form a dataset. Each dataset is then randomly split into a training subset (with 90\% of the samples) and a testing subset (with 10\% of the samples). For CNN-Cap which utilizes the grid-based data, we convert a sample structure to $n_c$ data where $n_c$ is the number of conductors in the sample. 
The number of grid cells ($L$) is set to 1024, which ensures that the grid size is less than the half of minimum spacing at any metal layer.
To avoid the interference of very small coupling capacitance, the coupling capacitance whose value is  less than 1\% of the corresponding total capacitance is not considered in the training stage and the prediction stage.

To tune the hyper-parameters for CNN-Cap, we randomly choose a layer combination under the 55nm technology to generate a dataset of Pattern-C. With it we tune the hyper-parameters to make the CNN-Cap's performance on the testing subset of this dataset best. The obtained hyper-parameters include batch size set to 64, learning rate set to $10^{-4}$ for predicting total capacitance and $10^{-5}$ for predicting coupling capacitance.

In the following subsections we will first evaluate the performance of CNN-Cap on Pattern-B structures, and then evaluate its performance on Pattern-C structures. Finally, we will study the effect of training subset's size and other evaluation metrics. 

\subsection{Results for Pattern-B Structures}
For Pattern-B structures, the MLP-based model presented in Section II.B can be used  as  the baseline. We call it MLP-Cap. Similar tuning is also applied to MLP-Cap, with a dataset of Pattern-B.  For the architecture, we tried different numbers of hidden layers and different numbers of neurons respectively. After some heuristic trials, we finally use an architecture with three hidden layers, and  256, 256 and 512 neurons are set in the three layers respectively.  Three nonlinear activation functions: ReLU, Sigmoid and Tanh are tested. The results show that Tanh function consistently surpass the other two. The grid search is used to find the appropriate learning rate and batch size. Finally, the batch size is set to $32$, and the learning rate is set to $10^{-5}$. Besides, the loss functions fo MSE (4) and MSRE (5) are tested with MLP-Cap. The results show that with the MSE loss function makes MLP-Cap perform better.

In this experiment, the pattern is the same as that in Fig. 3(b) which includes 1, 3 and 1 conductors on three metal layers respectively. And, the structure can be described with 9 geometric parameters, which are the input to MLP-Cap.

The results for predicting total capacitance and coupling capacitance, for eight datasets from the both technologies, are shown in Table \ref{table:B-total} and Table \ref{table:B-couple}, respectively. In the tables, every two rows correspond to a data set generated for a pattern. $Err_{avg}$ means the average of the relative error's absolute value. $Err_{max}$ means the maximum of the relative error's absolute value. Ratio(Err$>$5\%) in Table II means the ratio of the number of total capacitances with error larger than 5\% to the number of all total capacitances predicted. Ratio(Err$>$10\%) in Table III means the ratio of the number of coupling capacitances with error larger than 10\% to the number of all coupling capacitances predicted. Here, we regard  relative error of 5\% as the accuracy criterion for total capacitance, and the relative error of 10\% as the accuracy criterion for coupling capacitance.

The experimental results show that CNN-Cap always performs better than MLP-Cap. In more details, both CNN-Cap and MLP-Cap can predict total capacitance with less than $1\%$ relative error on average, while the maximum relative error of CNN-Cap is not larger than \textbf{1.3\%}. However, the maximum relative error of MLP-Cap is often larger than $5\%$. As for predicting coupling capacitance, the maximum relative error of MLP-Cap is always larger than 10\% and can be as large as 114\%. On the contrary, only a very few of coupling capacitances (at most \textbf{0.1\%} of all coupling capacitances) computed with CNN-Cap have relative error larger than $10\%$.
	\begin{table}[h]
	\setlength{\abovecaptionskip}{1pt}
	\centering
		\caption{DNN Models' Performance on Total Capacitance for Pattern-B}
		\label{table:B-total}
			\begin{tabular}{@{~}c@{~}c@{~}c@{~}c@{~}c@{~}c@{~}c@{~}c@{~}c@{~}c@{~}c@{~}}
				\hline
			Tech. Node & & Layers & &	Method & & $Err_{avg}$ & & $Err_{max}$ & & Ratio(Err$>$5\%) \\
				\hline
			55nm & & (2, 3, 6) & &	MLP-Cap & & 0.8\% & & 8.1\% & & 0.4\% \\
				\hline
			55nm & & (2, 3, 6) & &	CNN-Cap & & 0.2\% & & 1.2\% & & 0 \\
				\hline
			55nm & & (2, 4, 6) & &	MLP-Cap & & 1.2\% & & 11.4\% & & 0.7\% \\
				\hline
			55nm & & (2, 4, 6) & &	CNN-Cap & & 0.2\% & & 1.0\% & & 0 \\
				\hline
			15nm & & (1, 3, 5) & &	MLP-Cap & & 0.5\% & & 3.8\% & & 0 \\
				\hline
			15nm & & (1, 3, 5) & &	CNN-Cap & & 0.2\% & & 1.3\% & & 0 \\
				\hline
			15nm & & (1, 3, 8) & &	MLP-Cap & & 0.5\% & & 5.2\% & & 0.0\% \\
				\hline
			15nm & & (1, 3, 8) & &	CNN-Cap & & 0.2\% & & 1.1\% & & 0 \\
				\hline
			\end{tabular}
	\end{table}

	\begin{table}[h]
	\setlength{\abovecaptionskip}{1pt}
	\centering
		\caption{DNN Models' Performance on Coupling Capacitance for Pattern-B}
		\label{table:B-couple}
			\begin{tabular}{@{~}c@{~}c@{~}c@{~}c@{~}c@{~}c@{~}c@{~}c@{~}c@{~}c@{~}c@{~}}
				\hline
			Tech. Node & & Layers & &	Method & & $Err_{avg}$ & & $Err_{max}$ & & Ratio(Err$>$10\%) \\
				\hline
			55nm & & (2, 3, 6) & &	MLP-Cap & & 3.3\% & & 114\% & & 6.6\% \\
				\hline
			55nm & & (2, 3, 6) & &	CNN-Cap & & 2.4\% & & 13.5\% & & 0.1\% \\
				\hline
			55nm & & (2, 4, 6) & &	MLP-Cap & & 2.5\% & & 89.2\% & & 4.4\% \\
				\hline
			55nm & & (2, 4, 6) & &	CNN-Cap & & 1.4\% & & 11.8\% & & 0.0\% \\
				\hline
			15nm & & (1, 3, 5) & &	MLP-Cap & & 2.5\% & & 87\% & & 3.7\% \\
				\hline
			15nm & & (1, 3, 5) & &	CNN-Cap & & 1.1\% & & 9.6\% & & 0 \\
				\hline
			15nm & & (1, 3, 8) & &	MLP-Cap & & 2.0\% & & 49.0\% & & 1.4\% \\
				\hline
			15nm & & (1, 3, 8) & &	CNN-Cap & & 0.9\% & & 14.3\% & & 0.1\% \\
				\hline
			\end{tabular}
	\end{table}

\begin{figure}[h]
\setlength{\abovecaptionskip}{1pt}
  \centering
    \subfigure[]
    {\includegraphics[width=1.7in]{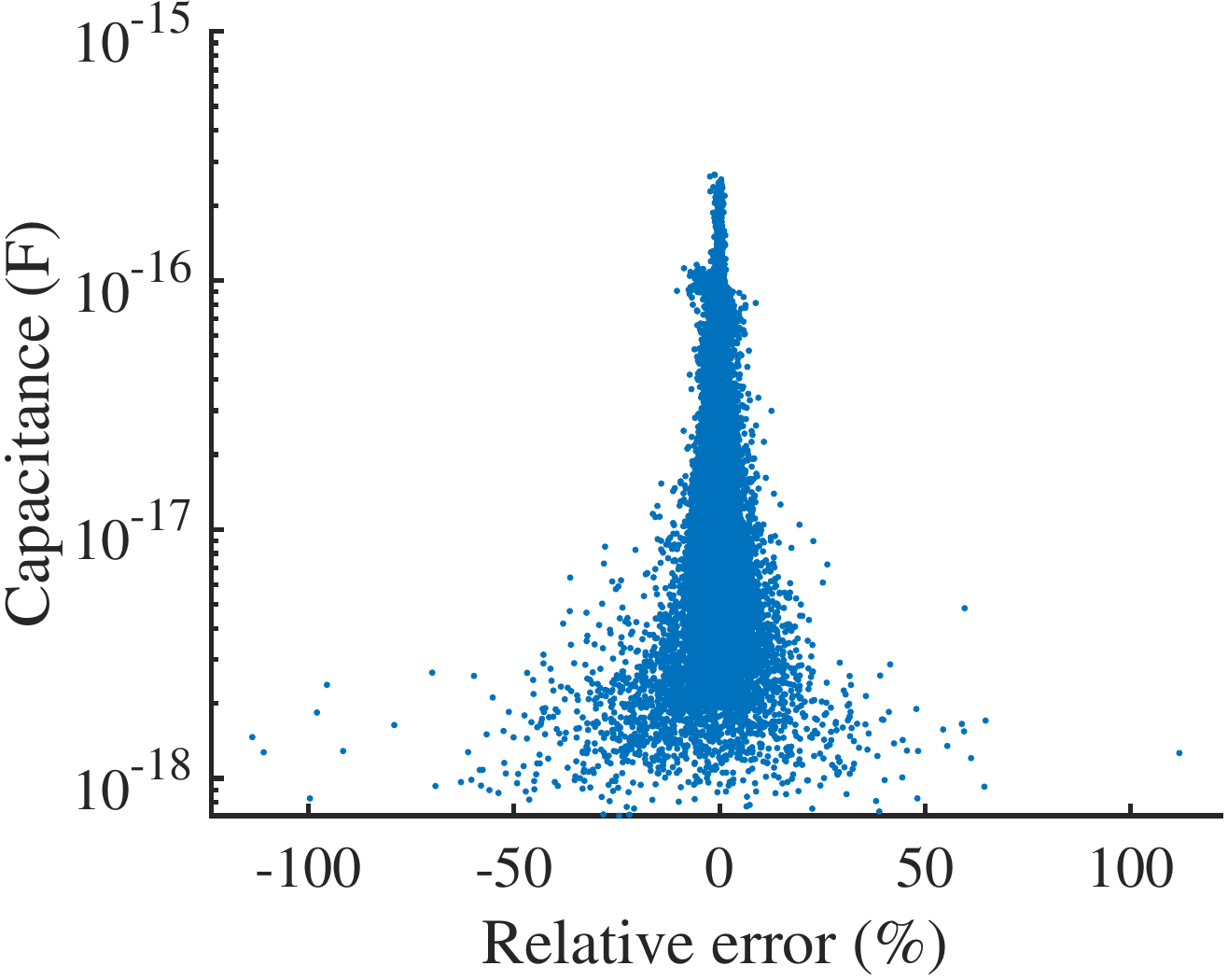}\label{fig:MLP_B}}
    \subfigure[]
    {\includegraphics[width=1.7in]{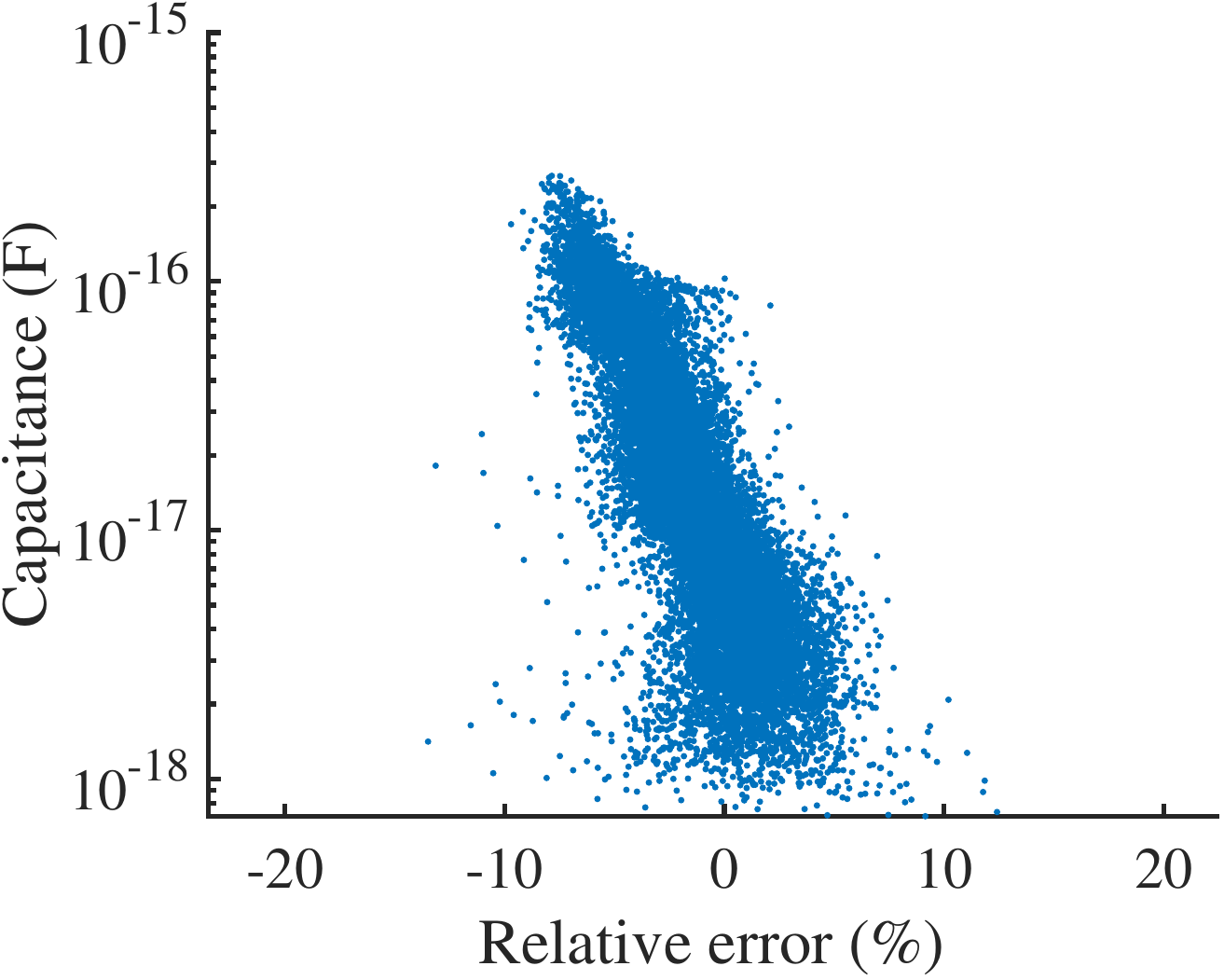}\label{fig:CNN_B}}  
  \caption{The calculated coupling capacitance versus relative error for Pattern-B. The dataset corresponds $(2,3,6)$ layer combination in the 55nm technology. (a) Results of MLP-Cap. (b) Results of CNN-Cap. }
      \label{fig:B-couple}
\end{figure}

Fig. \ref{fig:B-couple} shows the coupling capacitances calculated with MLP-Cap and CNN-Caps respectively, along the  relative error of each capacitance. From the figure we see  that CNN-Cap can accurately predict most coupling capacitance, and the few examples of inaccurate prediction usually corresponds to a small coupling capacitance. This means reasonable and acceptable accuracy. On the contrary, the MLP-Cap is unsatisfactory in many cases, and its maximum error is very large as shown in Fig. \ref{fig:MLP_B}. 

More dataset for different layer combinations have been generated and tested. The results show that for all the tested datasets, the average relative error of CNN-Cap on total capacitance and coupling capacitance are just 0.22\% and 1.36\%, respectively. This experiment demonstrates the good accuracy of CNN-Cap for Pattern-B structures. 

Another capacitance modeling approach which may have good accuracy is based on look-up table and the bilinear interpolation. For the considered Pattern-B structures, if each parameter takes 20 sample values the total numbers stored in the look-up table would be $20^9 = 5.12\times 10^{11}$, which leads to unacceptable memory cost in reality. This demonstrates the advantage of capacitance modeling technique based on neural networks. 

\subsection{Results for Pattern-C Structures}
For generating the sample structures of Pattern-C, we assume the total number of conductors on the up and down metal layers ranges from 6 to 8. Such a pattern with a variable number of conductors cannot be handled by a single MLP-Cap. So, we consider the hybrid method discussed at the end of Section III as the baseline. It is called  Grid+MLP model. It utilizes the grid-based data representation and shares the same architecture and hyper-parameters as MLP-Cap. And, it is trained with the same approach for MLP-Cap.

The results of CNN-Cap and Grid+MLP model for four datasets of Pattern-C  are listed in Table \ref{table:C-total} and Table \ref{table:C-couple}. The results show that CNN-Cap performs much better than Grid+MLP. The maximum relative error of CNN-Cap on total capacitance is not larger than \textbf{1.2\%}. As for coupling capacitance, the performance of the Grid+MLP is much worse and unacceptable, with the maximum error as large as 542\%. 
On the contrary, the relative error derived from the CNN-Cap is less than $10\%$ for more than \textbf{99.5\%} predicted capacitances. 
	\begin{table}[h]
	\setlength{\abovecaptionskip}{1pt}
	\centering
		\caption{DNN Models' Performance on Total Capacitance for Pattern-C}
		\label{table:C-total}
			\begin{tabular}{@{~}c@{~}c@{~}c@{~}c@{~}c@{~}c@{~}c@{~}c@{~}c@{~}c@{~}c@{~}}
				\hline
			Tech. Node & & Layers & &	Method & & $Err_{avg}$ & & $Err_{max}$ & & Ratio(Err$>$5\%) \\
				\hline
			55nm & & (2, 3, 6) & &	Grid+MLP & & 0.4\% & & 7.5\% & & 0.1\% \\
				\hline
			55nm & & (2, 3, 6) & &	CNN-Cap & & 0.1\% & & 1.0\% & & 0 \\
				\hline
			55nm & & (2, 4, 6) & &	Grid+MLP & & 0.7\% & & 7.7\% & & 0.1\% \\
				\hline
			55nm & & (2, 4, 6) & &	CNN-Cap & & 0.2\% & & 1.1\% & & 0 \\
				\hline
			15nm & & (1, 3, 5) & &	Grid+MLP & & 0.7\% & & 4.3\% & & 0 \\
				\hline
			15nm & & (1, 3, 5) & &	CNN-Cap & & 0.2\% & & 1.1\% & & 0 \\
				\hline
			15nm & & (1, 3, 8) & &	Grid+MLP & & 0.5\% & & 5.4\% & & 0.1\% \\
				\hline
			15nm & & (1, 3, 8) & &	CNN-Cap & & 0.2\% & & 1.2\% & & 0 \\
				\hline
			\end{tabular}
	\end{table}

	\begin{table}[h]
	\setlength{\abovecaptionskip}{1pt}
	\centering
		\caption{DNN Models' Performance on Coupling Capacitance for Pattern-C}
		\label{table:C-couple}
			\begin{tabular}{@{~}c@{~}c@{~}c@{~}c@{~}c@{~}c@{~}c@{~}c@{~}c@{~}c@{~}c@{~}}
				\hline
			Tech. Node & & Layers & &	Method & & $Err_{avg}$ & & $Err_{max}$ & & Ratio(Err$>$10\%) \\
				\hline
			55nm & & (2, 3, 6) & &	Grid+MLP & & 10.4\% & & 542.4\% & & 27.1\% \\
				\hline
			55nm & & (2, 3, 6) & &	CNN-Cap & & 1.2\% & & 38.6\% & & 0.3\% \\
				\hline
			55nm & & (2, 4, 6) & &	Grid+MLP & & 9.1\% & & 489.3   \% & & 22.7\% \\
				\hline
			55nm & & (2, 4, 6) & &	CNN-Cap & & 1.2\% & & 14.4\% & & 0.1\% \\
				\hline
			15nm & & (1, 3, 5) & &	Grid+MLP & & 9.8\% & & 492.8\% & & 24.8\% \\
				\hline
			15nm & & (1, 3, 5) & &	CNN-Cap & & 1.8\% & & 38.8\% & & 0.4\% \\
				\hline
			15nm & & (1, 3, 8) & &	Grid+MLP & & 11.4\% & & 390.4\% & & 30.8\% \\
				\hline
			15nm & & (1, 3, 8) & &	CNN-Cap & & 1.5\% & & 12.4\% & & 0.0\% \\
				\hline
			\end{tabular}
	\end{table}

Fig. 12 shows the calculated coupling capacitance (by CNN-Cap) versus relative error for Pattern-C structures. It reveals again that CNN-Cap has very good accuracy on coupling capacitance for most structures. For example, for the test data corresponding to layer combination (2, 3, 6) in the 55nm technology, only 0.3\% of all coupling capacitances has error larger than 10\%.  
\begin{figure}[h]
\setlength{\abovecaptionskip}{1pt}
  \centering
    \subfigure[]
    {\includegraphics[width=1.7in]{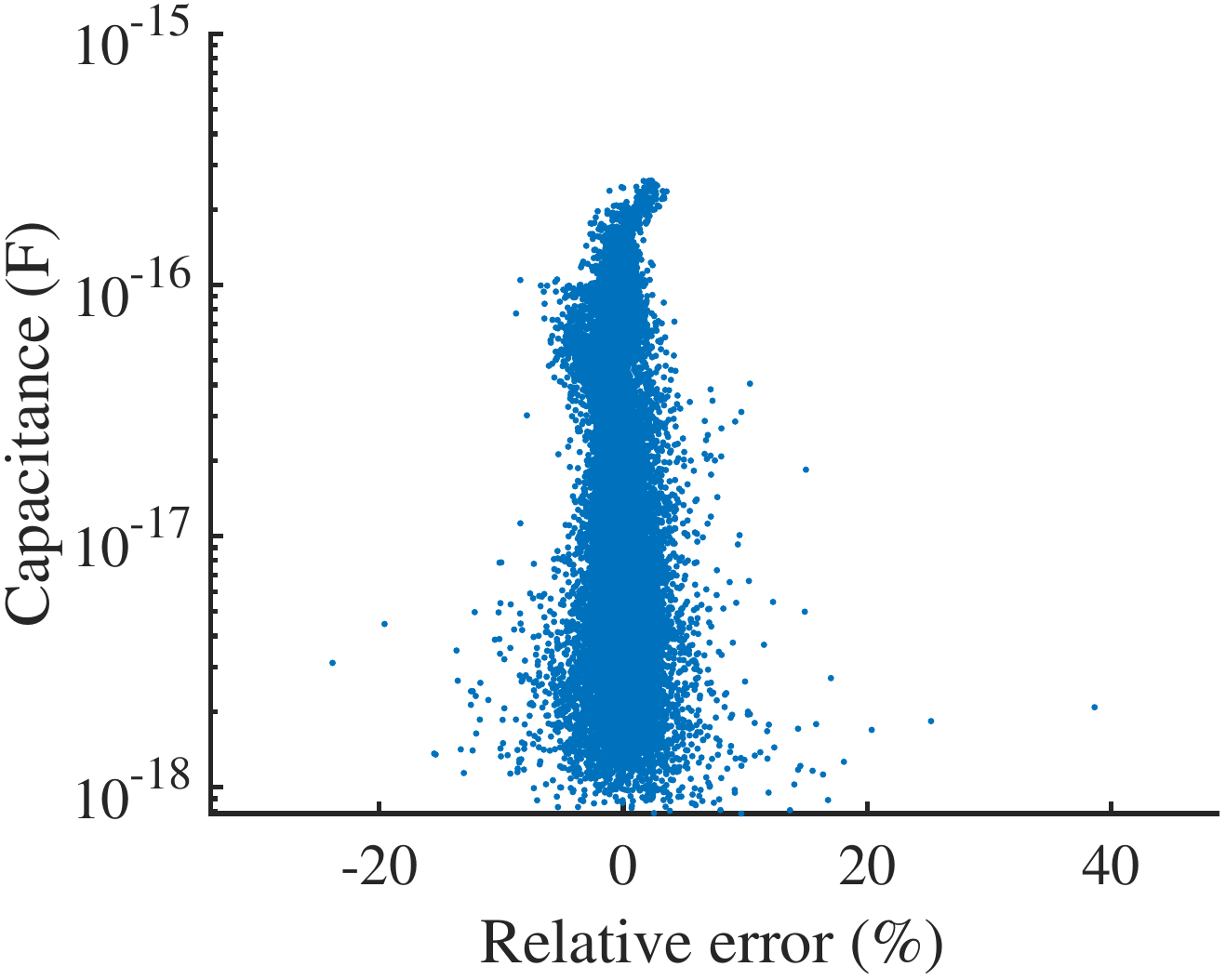}\label{fig:CNN0}}
    \subfigure[]
    {\includegraphics[width=1.7in]{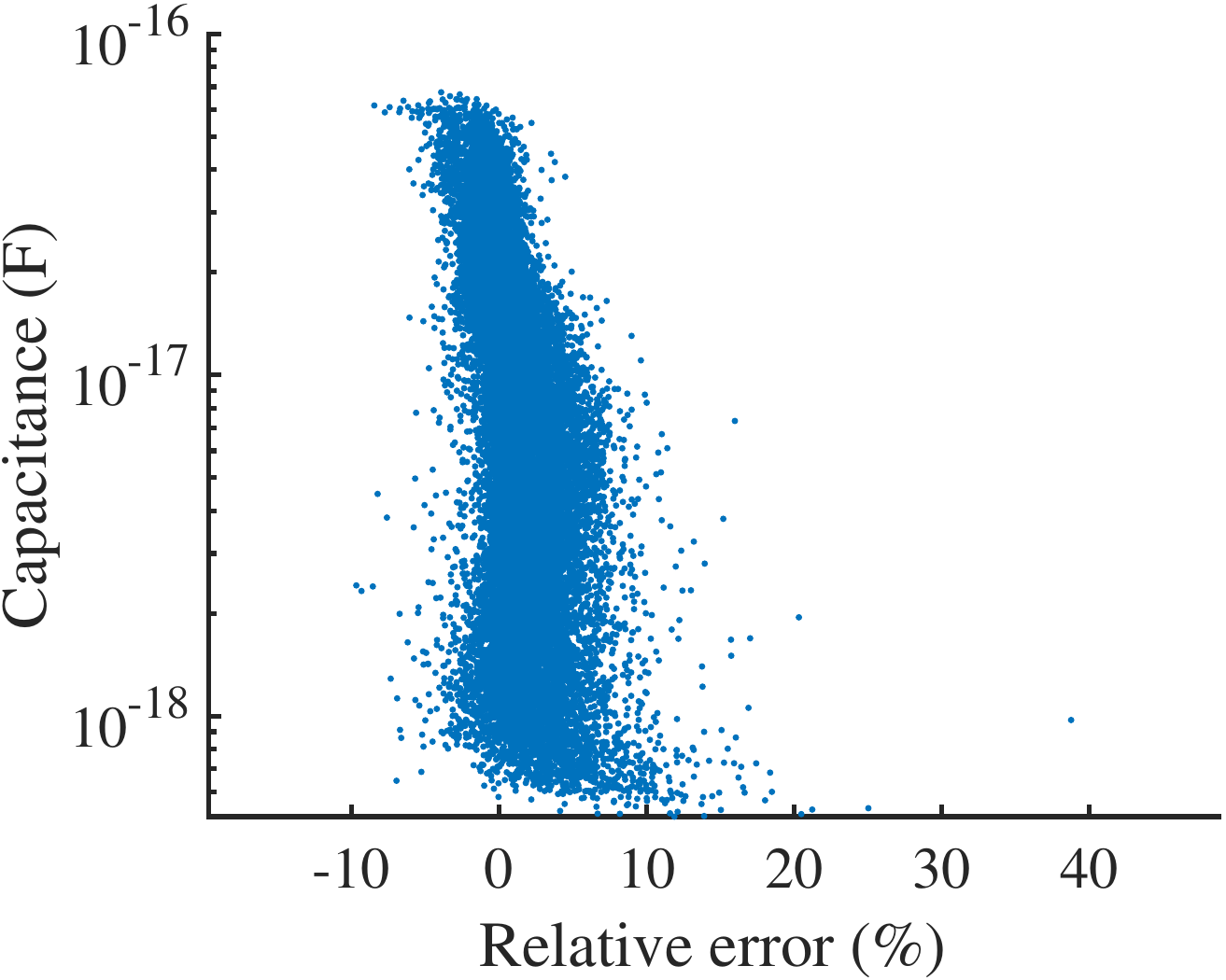}\label{fig:CNN1}}  
    \label{fig:C-couple}
  \caption{The calculated coupling capacitance versus relative error for Pattern-C. (a) Results of CNN-Cap for layer combination $(2,3,6)$ in the 55nm technology. (b) Results of CNN-Cap for layer combination $(1,3,5)$ in the 15nm technology. }
\end{figure}

More dataset for different layer combinations have been generated and tested. The results show that, for all the tested datasets the average relative error of CNN-Cap on total capacitance and coupling capacitance are just 0.18\% and 1.38\%, respectively. This experiment demonstrates the good accuracy of CNN-Cap for Pattern-C structures.

\subsection{The Effect of Training Set's Size and More Comparisons}
\begin{figure}[h]
\setlength{\abovecaptionskip}{1pt}
 \centering
    \includegraphics[width=2.8in]{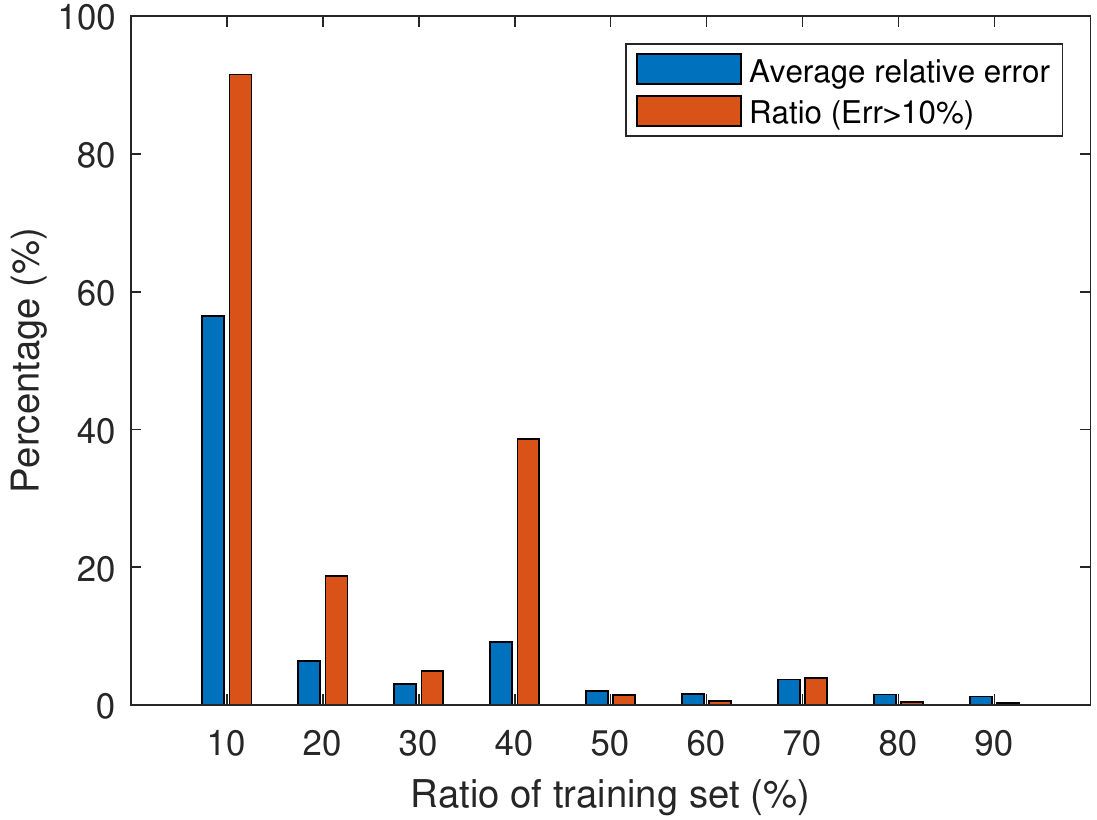}
 \caption{The accuracy of CNN-Cap on coupling capacitance vs. the size of training subset.}
    \label{fig:ratio_bar}
\end{figure}

An additional experiment is carried out to evaluate the effect of training set's size on CNN-Cap's accuracy. We change the ratio of the training subset from 90\% to 80\%, 70\%, down to 10\%, and then rerun the training process for a dataset of Pattern-C, respectively. For each resulted CNN-Cap model, we examine it with the  testing subset. The average relative error on coupling capacitance and the ratio of the coupling capacitances with error larger than 10\% are plotted in Fig. \ref{fig:ratio_bar}.  From it we see that, when the ratio of training subset is 50\% or larger (meaning with 25000 data or more) the trained CNN-Cap model always has fairly good accuracy. Similar results are observed for the total capacitance. Therefore, a training set with 25000 data through 45000 data is usually sufficient for building an effective CNN-Cap model.

We compare the prediction time of CNN-Cap with Raphael, i.e. the 2-D field solver \texttt{rc2} within Raphael. For a dataset  including 50000 Pattern-C structures, we extract the capacitances of them with CNN-Cap and Raphael respectively. Their average runtimes  per structure are listed in Table VI. From the table we see that CNN-Cap runs \textbf{4693X} faster than Raphael. Even though we can run multiple processes of Raphael on the machine, say 32 processes, calculating capacitance with CNN-Cap can still be \textbf{147X} faster than running the 2-D field solver.
\begin{table}[h]
	\setlength{\abovecaptionskip}{2pt}
\centering
	\caption{The Average Runtimes of CNN-Cap and Raphael for Calculating the Capacitances of a Pattern-C Structure}
	\label{table:time}
		\begin{tabular}{ccc}
			\hline
			Method & Time (ms) & Speedup\\
			\hline
			Raphael & 704 & --\\
			\hline
			CNN-Cap & 0.15 & 4693X\\
			\hline
		\end{tabular}
\end{table}

As for the model size, a CNN-Cap model has 14473418 parameters occupying 13.8 MB storage, which is much smaller than the look-up table based model. 
The training time for a CNN-Cap model is about 1.3 hours. Considering various layer combinations for a given process technology with 10 metal layers (like that in FreePDK15), we see that building all CNN-Cap models can be completed within a week on a GPU server with 8 Nvidia RTX2080Ti GPUs. And, the CNN-Cap models deliver much better accuracy than the existing methods for pattern capacitance modeling.

\section{Conclusions}
In this work, a CNN based capacitance model called CNN-Cap and the corresponding model-building techniques are proposed. 
They include a grid-based data representation for 2-D pattern structures, and a ResNet-like CNN architecture and corresponding training approach. CNN-Cap is able to predict the total capacitance and coupling capacitances for the 2-D pattern with a variable number of conductors. This largely reduces the number of patterns and the corresponding capacitance models employed in the pattern matching based full-chip capacitance extraction. Extensive experiments have demonstrated the advantages of CNN-Cap over MLP based models and traditional model-building approaches.

In the future, the proposed method may be extended to build capacitance models for 3-D interconnect structures. 

\bibliographystyle{IEEEtran}
\bibliography{main6final}

\end{document}